\documentclass[3p,12pt, authoryear,english]{elsarticle}

\usepackage[english]{babel}
\usepackage[utf8]{inputenc}
\usepackage{graphicx}
\usepackage{url}
\usepackage{datetime}
\usepackage{lineno}
\usepackage{amsmath}
\usepackage{todonotes}
\usepackage{booktabs}
\usepackage{algorithm} 
\usepackage{algpseudocode}
\usepackage{hyperref}
\usepackage{cleveref}
\usepackage{amssymb}
\usepackage{multirow}
\usepackage{bm}

\newcommand{\E}[2]{ {\mathbb E}_{#1} \left[ #2 \right] }
\newcommand{\V}[2]{ \mathbb{V}_{#1} \left[ #2 \right] }
\newcommand{\Vhat}[2]{ \hat{\mathbb{V}}_{#1} \left[ #2 \right] }
\newcommand{\vbar}{\,|\, }

\begin{document}

\begin{frontmatter}
	
\title{Scalable Probabilistic Forecasting in Retail with Gradient Boosted Trees: A Practitioner's Approach}

\begin{abstract}
The recent M5 competition has advanced the state-of-the-art in retail forecasting. However, we notice important differences between the competition challenge and the challenges we face in a large e-commerce company. The datasets in our scenario are larger (hundreds of thousands of time series), and e-commerce can afford to have a larger assortment than brick-and-mortar retailers, leading to more intermittent data. 
To scale to larger dataset sizes with feasible computational effort, firstly, we investigate a two-layer hierarchy and propose a top-down approach to forecasting at an aggregated level with less amount of series and intermittency, and then disaggregating to obtain the decision-level forecasts. Probabilistic forecasts are generated under distributional assumptions. Secondly, direct training at the lower level with subsamples can also be an alternative way of scaling. Performance of modelling with subsets is evaluated with the main dataset.
Apart from a proprietary dataset, the proposed scalable methods are evaluated using the Favorita dataset and the M5 dataset. We are able to show the differences in characteristics of the e-commerce and brick-and-mortar retail datasets. Notably, our top-down forecasting framework enters the top 50 of the original M5 competition, even with models trained at a higher level under a much simpler setting.
\end{abstract}

\begin{keyword}
probabilistic forecasting, gradient boosted trees, global models, disaggregation
\end{keyword}

\author[1]{Xueying Long}
\author[1]{Quang Bui}
\author[2]{Grady Oktavian}
\author[1]{Daniel F. Schmidt}
\author[1,3]{Christoph Bergmeir \corref{cor1}}
\author[1]{Rakshitha Godahewa}
\author[4]{Seong Per Lee}
\author[4]{Kaifeng Zhao}
\author[4]{Paul Condylis}

\address[1]{Department of Data Science and Artificial Intelligence\\
    Monash University, Australia.}
\address[2]{Data Science, Tokopedia, Indonesia}
\address[3]{Department of Computer Science and Artificial Intelligence\\
	University of Granada, Spain.}
\address[4]{Data Science, Tokopedia, Singapore}
\cortext[cor1]{Corresponding Author: bergmeir@ugr.es}

\end{frontmatter}

\section{Introduction}

Forecasting plays an important role in decision-making processes. In the retail industry, accurate sales forecasting is crucial for different phases such as supply chain management~\citep{babai2022demand} and inventory control~\citep{trapero2019quantile}. Furthermore, probabilistic forecasts are often more essential in these cases, which quantify the future uncertainties that should normally be considered, for example, for determining the stock level and reorder point. However, it is a challenging problem, among other things due to the fact that the series are often intermittent, i.e., a large percentage of entries is zero. 

The recent M5 competition~\citep{makridakis2021uncertainty} has established the state of the art of retail forecasting, both under an accuracy and an uncertainty track, that focused on probabilistic forecasting.
However, we observe that our use cases in a large Indonesian e-commerce retail company differ from the challenge posed in the M5 competition, which makes it necessary to evolve the methods accordingly. We focus on probabilistic forecasts as these are required in our use cases. Then, the two biggest differences we have identified are that in our practice, the datasets are often significantly larger and more intermittent than the dataset provided in the M5. While the M5 had less than \@ 50,000 time series, over half a million different types of products are purchased on our e-platform each day. Furthermore, the M5 data are from a traditional brick-and-mortar retail situation, which has some important differences to e-commerce, most notably that e-commerce can typically afford to have a larger assortment, and many products with slow sales, which leads to even more intermittency in the data. Apart from these differences in the data, our aim is to develop a system that is ready for production use, and as such also has some constraints regarding robustness and execution time that the M5 participants didn't have in the same way.

Consequently, the main aim of our work is to adapt the best-performing M5 methodologies to address these problems. The M5 was dominated by global models~\citep{januschowski2020criteria}, which are learned across series. This has important consequences for scalability, as global models cannot be fitted in parallel as trivially as local models, which are embarrassingly parallelisable along the series dimension. We need the methods to scale to datasets that are at least an order of magnitude larger than the M5 data. An intuitive way of scaling global models is to train the global models not in a truly ``global'' way, i.e., across all available series, but to train several such models on subsets of the data. This is a popular processing step, and most competitors in the M5 subdivided the data in one way or another. One of the earliest works proposing this procedure that we are aware of is the work of \cite{Bandara2020grouping}, and later this idea was studied more systematically in \cite{Godahewa2021localised}. However, subdividing the data has important modelling consequences and cannot be seen as a step with the sole purpose to achieve scalability. An intuitive alternative is to use simpler models, but this does not guarantee the ability to train models with a feasible computational effort, let alone the forecast accuracy. Another option is to train with less data. Straightforwardly, one can leave out part of the historical data and fit a model with subsamples. Apart from that, given the hierarchical structure of the data, we can also train models on a higher level with much less series and intermittency, and apply a top-down disaggregation to obtain lower level forecasts. All of these need to be done in a probabilistic way, and parametric methods with distributional assumptions are relatively simple and practical for implementations.

In the M5, tree-based methods were very successful and most competitors developed their solutions using LightGBM~\citep{ke2017lightgbm}, a highly efficient gradient boosted tree (GBT) algorithm. For example, the winning method in the accuracy track leveraged LightGBM by training on grouped data from multiple categories and combining the forecasts with equal weights~\citep{makridakis2022accuracy}. Tree-based implementations such as LightGBM and XGBoost~\citep{Chen2016XGBoost} are open source and flexible under different problem settings. As LightGBM offers fast training while maintaining predictive accuracy, it is currently a superior solution over other implementations of GBTs that produce less accuracy with longer training time. 

In this paper, we propose two efficient ways of generating accurate and scalable forecasting systems. We firstly make the most of a two-layer hierarchy of raw and aggregated data, and develop a top-down forecasting framework that is able to scalably predict with small computational effort while maintaining competitive accuracy. Instead of directly dealing with data on the decision level, we forecast with the aggregated series and disaggregate back in a top-down fashion according to historical proportions. Our forecasting framework is capable of generating accurate probabilistic forecasts with simple assumptions of distributions. Besides, we utilise subsamples of the data to enhance the computational performance and build models based on linear regression. Such models are able to capture the overall characteristics of the large amount of data and produce reliable forecasts. The two proposed approaches are analysed on our proprietary e-commerce dataset, as well as the public Favorita dataset and the M5 competition dataset. As a notable side-product of this research, we have implemented a negative binomial loss function for LightGBM~\citep{ke2017lightgbm}, for which the details are given in~\ref{appendix:a}.

The rest of this paper is organised as follows. Section~\ref{sec:related-work} reviews the related work. Section~\ref{sec:methodology} provides a comprehensive description of the proposed top-down forecasting framework and the subsampling procedure. Section~\ref{sec:experimental-framework}  explains the experimental setup. Section~\ref{sec:evaluation} reports the results and provides a further discussion. Section~\ref{sec:conclusion} concludes our work. 

\section{Related work}
\label{sec:related-work}
In this section, we cover the prior work in the relevant areas of this research, which are global models, hierarchical-, probabilistic-, and intermittent forecasting.

\subsection{Modelling across series with global models}

Global modelling~\citep{januschowski2020criteria} has received a lot of attention lately in the forecasting community, and all top contenders in the M5 are global models. Under this paradigm, a single model is built across many series, with shared parameters. As a global model is trained with more data, it can afford to be more complex, compared with local per-series models. \cite{MonteroManso2021principles} present some theoretical explanations for the superiority of global models over local models, and argue that no similarity or relatedness between series is necessary for global models to work well. \cite{hewamalage2022global} confirm these findings empirically and make them more nuanced in a simulation study, arguing that no assumptions on relations of time series are necessary beforehand as global models have the capacity of learning complex patterns and perform well even when the series are heterogeneous.
One of the earliest and most prominent global models in the literature is DeepAR~\citep{salinas2020deepar}, which is a global forecasting method with autoregressive neural networks. It demonstrates high forecasting accuracy for Amazon sales data and can be considered a standard benchmark in retail forecasting. Also, global models have shown huge success in Kaggle competitions over the years~\citep{bojer2021kaggle}. 
Consequently, we focus in our research on global models, as research has well established by now their superiority over local models in a retail setting like ours.

\subsection{Hierarchical forecasting}

Retail sales data is naturally organised in a hierarchy, i.e., per-store product sales data at the bottom level can be added up according to product categories and regions. Typically, hierarchical forecasting is concerned with producing coherent forecasts across different levels of the hierarchy (for different decisions to be made, such as strategical, tactical, or operational decisions). Furthermore, hierarchical forecasting methods have been used in the past to transport information between series, such as bringing seasonal patterns only emerging at higher levels of the hierarchy into the noisy bottom-level series forecasting.
Classical approaches of hierarchical forecasting in the literature are top-down, bottom-up and middle-out methods~\citep{hyndman2011optimal}, where forecasts are produced on only a single level of the hierarchy and then aggregated up or disaggregated down, using historical (or through other ways obtained) proportions. More sophisticated alternatives are optimal reconciliation approaches~\citep{hyndman2011optimal}, where all series in the hierarchy are forecast, and then in a subsequent step a reconciliation (optimisation) is performed to adjust the forecasts and make them coherent. The most recent methods combine forecasting and reconciliation into a single step, building global models that are able to produce reconciled forecasts directly. The most prominent methods in this space are~ \cite{Rangapuram2021HierE2E, han2021simultaneously, paria2021hierarchically, kamarthi2022profhit}.

Our motivation for using a hierarchy is different to the usual use cases. We leverage the hierarchical structure not from the perspective of reconciliation, and are not interested in coherent forecasts for the whole hierarchy. Instead, we use the hierarchy as a way to scale the forecasts from more aggregated levels in the hierarchy, where fewer time series exist, to lower levels where the amount of series and their intermittency hinder traditional forecasting techniques. Thus, the sophisticated methods from the literature are not directly applicable to our use case.

\subsection{Probabilistic forecasting for intermittent data}

Forecasting expectations is important, though, predicting the distribution can explicitly provide information on the uncertainty of the produced forecasts, thus being more beneficial for decision-makers. We categorise the existing probabilistic forecasting approaches into two main parts: non-parametric methods such as quantile regression and bootstrapping, and parametric methods which predict the parameters under some distributional assumptions. With a certain loss function (i.e., pinball loss function), quantile forecasts can be directly generated, and the implementation is available in most open-source GBT frameworks. However, the modelling and training process needs to be repeated for each quantile of interest. Regarding probabilistic forecasting for intermittent data, \cite{lainder2022forecasting} propose a quantile forecasting method with LightGBM and data augmentation techniques, which won the first place of the M5 uncertainty track. \cite{willemain2004approach} propose a bootstrapping method with a two-state Markov model for intermittent demand, which leads to more accurate results than exponential smoothing and Croston's method. \cite{viswanathan2008boostrapping} then improve this method by generating demand intervals according to the historical distribution separately. Even with some highlights in forecast accuracy, bootstrapping requires a large amount of historical data and huge computational costs, which poses a question on its efficiency and necessity in real cases~\citep{syntetos2015forecasting}.

On the other hand, parametric methods involve understanding the characteristics of historical data and the data generating process. Classical choices of fitting retail data in the literature are the use of a Poisson distribution~\citep{heinen2003modelling, Snyder2012Forecasting}, or a negative binomial distribution~\citep{agrawal1996estimating, Snyder2012Forecasting}, mixed with zero-inflated models~\citep{lambert1992zero} and hurdle models~\citep{Cragg1971Statistical} to accommodate to the excess zeros. Based on the distributional assumptions, parameters are then learned empirically or through some algorithms. \cite{Snyder2012Forecasting} propose a hurdle shifted Poisson model and introduce a dynamic state-space structure in both damped and undamped versions. Following that, ~\cite{Jiang2019approach} use the mixed zero-truncated Poisson hurdle model and find better forecast accuracy. \cite{deRezende2021ISSM} extend the state-space structure with the negative binomial distribution of~\cite{Snyder2012Forecasting} by considering the external seasonal and causal factors, and their solution achieved the sixth place of the M5 uncertainty competition. Parameter estimation of such state-space models is often obtained via the maximum likelihood method or the expectation maximisation (EM) algorithm, which can be computationally intensive. \cite{kolassa2016evaluating} studies a set of parametric methods with Poisson and negative binomial assumptions, and further emphasises the consideration of over-dispersion in retail data. However, that work only focused on local methods, and did not consider ways of scaling up the forecasting process.

Unlike many machine learning algorithms which are only capable of producing a single output, the Generalised Additive Models Location Shape Scale~\citep[GAMLSS, ][]{Stasinopoulos2007generalized} approach can generate all parameters of the assumed distributions at the same time. \cite{ziel2021m5} applies this approach to the M5 dataset with different distribution assumptions, including a zero-inflated Poisson distribution. A major pitfall of GAMLSS is the huge computational cost, for which models are trained only based on subsamples in that work. Following the literature, we consider the Poisson distribution and negative binomial distribution, as mixed distributions add on extra parameters which would introduce additional complexity during the modelling process. Moreover, these two distributions are characterised as being infinitely divisible~\citep{steutel2003infinite}, for example, a Poisson random variable can be expressed as the sum of an arbitrary number of i.i.d. Poisson random variables. In this case, we can decompose the aggregated level forecasts and generate probabilistic forecasts based on the distributions for both layers. \cite{Olivares2021poissonmixture} have tested the Poisson mixtures from a perspective of hierarchical reconciliation while modelling with a deep neural network. We also examine the negative binomial mixtures as an extra dispersion parameter could introduce more variability in modelling.

\section{Methodology}
\label{sec:methodology}

As outlined earlier, our methodology consists of improvements of the retail forecasting state of the art in different dimensions, to address larger amounts of data and intermittency. 
In particular, we propose a methodology consisting of: (1) a data partitioning step commonly used in retail settings, with details in Section~\ref{sec:demand-classification}, and (2) a hierarchical top-down approach, where we have too many series on the bottom level to forecast them directly, and therefore we resort to forecasting on a higher aggregation level and then de-aggregate the forecasts, in Section~\ref{sec:top-down}. Finally, we (3) use a simple sub-sampling approach to train on only a sub-sample of the dataset to make the training task more tractable, in Section~\ref{sec:sampling}. 

\subsection{Demand classification}
\label{sec:demand-classification}

Following the scheme proposed by~\cite{Syntetos2005categorization}, we classify the series first by variance in demand timing and quantity into four groups: smooth, erratic, lumpy, and intermittent, according to the Average Demand Interval (ADI) and Coefficient of Variation squared ($\textrm{CV}^2$) as defined in Equation~\eqref{eq:ADI_CV2}, with threshold values of 1.32 and 0.49 respectively. Even though these threshold values were originally proposed for selecting an optimal method between simple exponential smoothing (SES) and a modified Croston's method~\citep{syntetos2005accuracy}, methods which we are not using in our work, we employ these threshold values as they are well-established in the literature.

\begin{equation}\label{eq:ADI_CV2}
  \begin{aligned}
	\textrm{ADI} &= \frac{\textrm{Days available since first sale}}{\textrm{Days with sale}}, \\
	\textrm{CV}^2 &= \left(\frac{\textrm{Standard deviation of daily sales}}{\textrm{Mean of daily sales}}\right)^2.
  \end{aligned}
 \end{equation}

Partitioning series into these groups is reasonable as series in these groups have characteristics which make them behave quite differently both when fitting models and when evaluating their forecasts. Regarding fitting, different loss functions are adequate. Regarding evaluation, if we evaluate all series together, the smooth and erratic series are likely to dominate the error measure when using scaled metrics; likewise for a scale-free measure, the intermittent and lumpy series will typically lead to a large part of the error, and dominate any error measure computed. In practice, the series with high (percentage) error are usually the intermittent ones, which are also the least important for the business. Thus, we first perform a demand classification and then fit models and evaluate them using scaled metrics for each group separately.

\subsection{Top-down distributional forecasting framework}
\label{sec:top-down}

\begin{figure}[H]
\includegraphics[width=0.99\linewidth,]{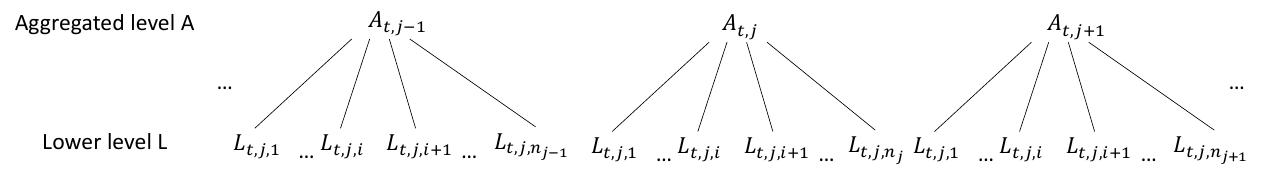} 
\caption{An illustration of the two-layer hierarchical structure.} 
\label{fig:top_down_structure}
\end{figure}

If we denote the series at the decision level as level L, we can then aggregate the series based on product hierarchy to an aggregated level, denoted as level A. The constructed two-layer hierarchy is illustrated in Figure~\ref{fig:top_down_structure}. At each time point $t$, a series $j$ at level A is denoted as $A_{t,j}$, and can be constructed by the sum of the corresponding $n_j$ series at level L. $L_{t,j,i}$ is used to denote a series $i$ at level L, where $j$ matches the $j$th series at level A in the hierarchy. We train global models at level A, a level in the hierarchy that is still a low level, but high enough that the data are less intermittent, and the number of series to be forecasted is feasible. We then produce forecasts recursively for the whole horizon. Any off-the-shelf global forecasting model can be used in this framework to generate point forecasts at level A; that is, the top-down distributional framework is model-agnostic. In this work, we implement it with LightGBM models and linear models. Considering a certain time point $t$ in the horizon $h$, we denote the point forecast at the aggregated level as $\hat{A}_{t,j}$ for series $j$ to be the estimated mean value.

In particular, the proposed forecasting framework consists of four steps: at each time point in the forecast horizon, we (1) point-forecast the values at the aggregated level A as predicted means, (2) estimate the parameter(s) of distribution at the aggregated level using the estimated mean, (3) obtain the historical proportion of lower-to-higher level sales, and disaggregate to obtain the lower level L point forecast, and (4) estimate the parameter(s) at the lower level. In this section, we start by introducing the distribution properties and then discuss each step in detail.

\subsubsection{Distribution properties and forecasting}
\label{sec:distribution-properties}

We assume that sales are Poisson random variables, or negative binomial random variables. For Poisson distributed sales $X \sim \rm Poisson(\lambda)$ with parameter $\lambda$, once we have estimated the mean, the parameter $\lambda$ can be estimated as the same value according to maximum likelihood estimation.
We can then estimate the distributional forecast by using the point forecast as an estimate of $\lambda_{A_{t,j}}$.
 
For the negative binomial distribution, considering a series of Bernoulli trials until we reach $r$ successful trials, when the probability of success in each trial equals to $p$, a random variable $X$ which denotes the number of failures follows a negative binomial distribution $X \sim \rm NB(r, p)$~\citep{RNegBin2021}, and the probability mass function is given by

$$P(x \vbar r,p) = {r+x-1\choose r-1}p^{r}(1-p)^{x}.$$
There is no closed form of maximum likelihood solution for the parameters, instead, we estimate the parameters through the method of moments as follows,
$$\hat{p} = \frac{\E{}{X}}{\V{}{X}} \; \; {\rm and} \; \;
\hat{r} = \E{}{X} \left( \frac{\hat{p}}{1-\hat{p}}\right). $$ 
Since $0 \leq p \leq 1$, the variance of the negative binomial distribution is greater than its mean, which is also known as over-dispersion. 
To estimate the distributional forecast, an estimated variance at each time point $t$ is necessary for parameter estimation. We treat the variance of sales over the training set $\Vhat{}{A_j}$ as a proper estimate, thus we get
$$\hat{p}_{A_{t,j}} = \frac{\hat{A}_{t,j}}{\Vhat{}{A_j}}\; \; {\rm and} \; \;
\hat{r}_{A_{t,j}} = \hat{A}_{t,j} \frac{\hat{p}_{A_{t,j}}}{1-\hat{p}_{A_{t,j}}}.$$

\subsubsection{Disaggregation}

We begin by forecasting the sales at the aggregated level, then apply a top-down approach to disaggregate the forecast to the lower level. In the hierarchy, the following property is satisfied at each time point $t$,
\begin{align*}
  A_{t,j}   &= \sum_{i=1}^{n_j}L_{t,j,i},
\end{align*}
where $A_{t,j}$ and $L_{t,j,i}$ respectively denote aggregated and lower level series in the hierarchy, and the number of series at the lower level equals to $n_j$ for a specific $A_{t,j}$ series at the aggregated level.

The disaggregation is performed by weighing the sales forecasts with the historical proportion of lower-to-higher sales over the training set. This proportion $\rho_{j,i}$ is calculated according to Equation~\eqref{eq:rho}, where $T$ is the timestamp of the last observation in the training set,

\begin{equation}\label{eq:rho}
  \rho_{j,i} = \frac{\sum^{T}_{t=1} L_{t,j,i}}{\sum^{T}_{t=1} A_{t,j}}.
\end{equation}
The point forecast at the lower level $\hat{L}_{t,j,i}$ is then given by $\hat{L}_{t,j,i} = \rho_{j,i}\hat{A}_{t,j}$.

\subsubsection{Parameter estimation for lower level series}

Similarly, with the Poisson assumption, a lower level point forecast $\hat{L}_{t,j,i}$ becomes the estimated mean of sales, which is also equal to the estimated $\lambda_{L_{t,j,i}}$.

Furthermore, the variance of the sales can be estimated through the in-sample variance at the lower level, denoted as $\Vhat{}{L_{j,i}}$. The estimation for the negative binomial distribution parameters at the lower level can then be conducted in a similar fashion,
$$\hat{p}_{L_{t,j,i}} = \frac{\hat{L}_{t,j}}{\Vhat{}{L_{j,i}}}\; \; {\rm and} \; \;
\hat{r}_{L_{t,j,i}} = \hat{L}_{t,j} \frac{\hat{p}_{L_{t,j,i}}}{1-\hat{p}_{L_{t,j,i}}}.$$
Moreover, it seems to be reasonable to assume that the value of parameter $p$ is fixed across the hierarchy, that is, series are independent and the value of $p$ is a constant. Then we would use the estimated $\hat p$ from the aggregated level ($\hat{p}_{A_{t,j}}$) as an estimate of $\hat p_{L_{t,j,i}}$ for the lower level series. However, in preliminary experiments not reported here, this procedure did not yield promising results and therefore we did not pursue this approach further.

\subsection{Sampling at the lower level}
\label{sec:sampling}

Another way of dealing with a situation where we have too much data to train on is a simple subsamplping approach, where we train the model on a subset of the data and use that model to forecast all series of interest. Also subsampling is model-agnostic, and can be employed with any method. However, as in our experiments we want to show how much accuracy we may potentially loose when training with subsamples compared with training on the full dataset, we need an approach that can also scale to large datasets properly. Thus, we use a simple linear regression model for this task, and solve the coefficients in matrix form for scalability. The input matrix $\mathbf{X}$ with a dimension of $n \times (p+1)$ can be defined as the original feature vectors augmented by a vector of ones at the first column, where $n$ is the number of observations and $p$ corresponds to the number of features. A coefficient vector ${\bm \beta}$ of the linear model is $(p+1)$ dimensional, with the first entry being the intercept. Then the residual sum-of-squares (RSS) can be defined as
$$\rm RSS = ||\mathbf{y} - \mathbf{X}{\bm \beta}||^2,$$
where $\mathbf{y}$ denotes the target vector, and $||{\bf x}||^2$ denotes the usual $\ell_2$ norm of the vector ${\bf x}$. We can obtain an estimate of coefficients by minimising the RSS, that is,
$$\hat{\bm{\beta}} = (\mathbf{X} ' \mathbf{X})^{-1} \mathbf{X}' \mathbf{y}.$$

In practice, we can first construct the matrix $\mathbf{X_i} ' \mathbf{X_i}$ and $\mathbf{X_i}' \mathbf{y_i}$ for each series $i$, which are $(p + 1) \times (p + 1)$ and $(p + 1) \times 1$ dimensional, respectively. This relaxes the memory requirements to a large extent. If we denote the sample size as $s$, a summation step is firstly applied,
$$\mathbf{X} ' \mathbf{X} = \sum_{i=1}^{s} \mathbf{X_i} ' \mathbf{X_i},$$
$$\mathbf{X}' \mathbf{y} = \sum_{i=1}^{s} \mathbf{X_i}' \mathbf{y_i}.$$
Then we solve for the coefficients. In these cases, we focus on point forecasting analysis as it is also sufficient to provide us with the information on the representativeness of the subsamples over the whole dataset.

\section{Experimental framework}
\label{sec:experimental-framework}

This section describes the datasets, benchmarks, and error measurements used in our experimental study.

\subsection{Datasets}

We are aware of two openly available large retail datasets, namely the M5 dataset and the Favorita dataset. Both these datasets represent traditional brick-and-mortar sales datasets. We use these datasets in addition to our proprietary e-commerce dataset. Based on demand classification, we can categorise the lower level series into four classes, and the percentage of series that fall into each class is summarised in Table~\ref{tab:data-summary}. We find that in our e-commerce dataset, lumpy and intermittent series are the biggest subgroups. The Favorita dataset contains series which are more evenly distributed over the four categories, while the intermittent series form a large part of the M5 dataset as well. We further calculate the percentage of zeros in each category of the three datasets. From Table~\ref{tab:data-summary}, our e-commerce series are more intermittent compared with the brick-and-mortar datasets we also use in the experiments. We describe the datasets in more details in the following.

\begin{table}[H]
  \centering
  \caption{Summary of the percentage of series and percentage of zeros across all series, in each category on the lower level of the three datasets analysed in this paper (in percent).}
    \begin{tabular}{rrrrrr}
    \toprule
          & Dataset & Smooth & Erratic & Lumpy & Intermittent \\
    \midrule
    \multirow{3}[0]{*}{Percentage of series} & E-commerce &  0.08 &  0.46&  39.53 &  59.93 \\
	  & Favorita & 20.51 & 19.58 & 33.62 & 26.29 \\
          & M5 & 6.23 & 2.83 & 18.38 & 72.56 \\
    \midrule
    \multirow{3}[0]{*}{Percentage of zeros} & E-commerce & 56.64 & 46.12 & 89.51 & 95.85 \\
	  & Favorita & 34.44 & 37.68 & 68.83 & 74.61 \\
          & M5 & 30.62 & 27.58 & 61.23 & 74.50 \\
   \bottomrule
    \end{tabular}
  \label{tab:data-summary}
\end{table}

\subsubsection{Our proprietary e-commerce dataset}

This dataset consists of series of daily sales across all regions of Indonesia from May 7th of 2019 to May 8th of 2021 from one particular department of the company. The total number of series is 246,361 on the lowest level. 

In the dataset, similar products are grouped and regarded as a `Catalogue',
and products in a catalogue have a high level of similarity in price. For example, an iPhone 11 could be one item of the catalogue, which contains different specific models such as green iPhone 11. We use the catalogue level as level A, and the specific models level as level L in the experiments.
To evaluate the top-down approach, we forecast 28 days ahead and evaluate the 10th and 90th percentile forecasts at level L. A further sampling analysis with point forecasting 28 days ahead is conducted with data from the lumpy and intermittent categories of this department.

\subsubsection{The Kaggle Favorita dataset}
The Favorita dataset~\citep{favoritadataset} provides daily unit sales data in brick-and-mortar grocery stores from January 1st of 2013 to August 15th of 2017. The original data contains negative values which denote the number of returns for a certain product, however, these negative values are set to zero in our experiments as we are only interested in sales forecasting. A natural way of constructing a two-layer hierarchy is to use the original data as the lower level, and sum up unit sales by item as an aggregated level, i.e., add up the volumes in different stores for each item. In this way, level A contains 3998 series, whereas level L consists of 172,906 series. 
The tasks performed are similar: we forecast the 10th and 90th percentiles of future 28 days ahead in a top-down fashion with models trained by the aggregated series as a whole, and analyse the performance of modelling on subsamples by demand class at level L.

\subsubsection{The M5 dataset}
With data available for over 5 years, participants were required in the original competition to submit 9 quantile forecasts for each series. The provided sales data is hierarchically structured and can be aggregated to 12 different levels. To give further insights of the proposed methods, we evaluate the performance of the proposed top-down probabilistic forecasting framework in line with the competition settings, where we utilise the hierarchy between level 10 (product unit sales aggregated by stores, 3049 series) and level 12 (product unit sales, 30,490 series, the lowest level). Models are trained with data from level 10 and forecasts are disaggregated proportionally to level 12, and quantile forecasts are then generated according to distributional assumptions.

\subsection{Compared settings}

The proposed top-down forecasting framework is implemented with LightGBM model variants and linear models variants. Models are trained with 100 lags as input features to capture possible weekly, monthly, and quarterly seasonality while being not too computationally expensive and complex. The LightGBM models are named by the corresponding loss functions and parameter settings, and linear models are named by specific regression settings. In the following, we list the techniques used in this work.

\begin{description}

\item[LightGBM] LightGBM models are trained in a top-down fashion under different loss functions and parameter settings. The LightGBM package provides L1, L2, Poisson, Huber, and Tweedie loss functions for regression problems~\citep{lightgbm-package-r}. It is also reasonable to consider over-dispersed loss functions such as negative binomial loss, however, as no off-the-shelf implementation of the negative binomial loss function is available, we implement it with the custom loss and evaluation function in Python (refer to \ref{appendix:a}). In terms of the parameter configurations, we consider default regression parameters, and a preset parameter setting used for the M5 competition~\citep{Bandara2021fast}, which are named as \verb|default| and \verb|preset| in the models, respectively. Instead of modelling with a constant, piecewise linear trees use linear functions to produce the outcomes, and have demonstrated accurate performance in  forecasting~\citep{Godahewa2022SETARTree}. So we also include the piecewise linear GBTs, which can be selected with the \verb|linear_tree| parameter in LightGBM. 

\item[Linear models] Ordinary least squares, or Pooled Regression~\citep[PR, ][]{Gelman2006pr} models linear relationships between predictors and target values. Penalised linear regression, specifically Lasso regression models~\citep{Tibshirani1996lasso} are also trained in the experiments. We implement penalised models with the R \verb|glmnet| package~\citep{glmnet2011package} under default settings with cross-validation. Moreover, apart from using the 100 lags as stated previously, it is intuitive to consider quadratic terms in the regression models. We trained models with Lasso penalty and extra 100 quadratic lag terms, however, they did not show improvements in accuracy so results are not reported here.

\item[GAM] Generalised Additive Models~\citep[GAM, ][]{Hastie1986gam} can be regarded as linear combinations of predictors after non-linear transformation. Similar to linear models, the effect of each predictor is independent, but transforming predictors through non-linear smooth functions allows models to fit more flexibly to the data while retaining much of the interpretability. We train GAM models with negative binomial regression and default parameters based on the R \verb|mgcv| package~\citep{mgcv2003package}.

\end{description}

In terms of benchmarks, we consider the following baselines of forecasts directly performed on level L, namely direct quantile modelling with LightGBM models, DeepAR, traditional univariate forecasting models, and in-sample quantiles. An input window of 100 lags is used for the former two approaches, similarly to the proposed methods. The details are as follows.

\begin{description}

\item[Direct LightGBM] Direct quantile models are trained on the lower level L to get the lower level prediction. This approach requires training a model for each quantile and each forecast step. We use LightGBM with the preset parameters from~\cite{Bandara2021fast} and the quantile loss function. 

\item[DeepAR] The autoregressive neural network forecasting framework developed by \cite{salinas2020deepar} is another competitive standard benchmark nowadays. We trained DeepAR models globally with the Python \verb|GluonTS| package~\citep{Alexandrov2020GluonTS} on the lower level L with default parameters and a negative binomial output. Considering the massive computational costs, we use DeepAR as a prototype for other deep-learning methods.

\item[Local statistical methods] Five classic statistical methods, namely Autoregressive Integrated Moving Average model~\citep[ARIMA,][]{box2015arima}, ExponenTial Smoothing model~\citep[ETS,][]{hyndman2008forecasting}, Mean, Naive, and Seasonal Naive (SNaive) are considered in the experiments. Models are fitted using the R \verb|fable| package~\citep{Hyndman2021fable} under their default configurations.

\item[In-sample quantiles]  If we take the distribution of the in-sample data as an estimate of the true marginal distribution, quantile forecasts in the future horizon can be then obtained according to this distribution, denoted as in-sample quantiles. The in-sample quantile forecasts on the lower level can be thought of as the probabilistic variant of a mean forecast for point forecasts. The in-sample quantiles of interest are calculated separately for each series. 

\end{description}

\subsection{Evaluation metrics}
Following the setup of the M5 competition, we evaluate the probabilistic forecasts using the Weighted Scaled Pinball Loss~\citep[WSPL, ][]{makridakis2021uncertainty}. We denote $q_t^{[u]}$ as the predicted value for quantile $u$ at time $t$, and $y_t$ as the corresponding ground truth. Then, for a series $i$, the Scaled Pinball Loss (SPL) is calculated for each quantile as follows,
$${\rm SPL}_i [u] = \frac{1}{h}\frac{\sum_{t=T+1}^{T+h}( u(y_t - q_t^{[u]})\mathbf{1}\{q_t^{[u]} \le y_t \} + (1-u)(q_t^{[u]} - y_t)\mathbf{1}\{q_t^{[u]} > y_t \} )}{\frac{1}{n-1}\sum_{t=2}^{T}|y_t - y_{t-1}|},$$
where the pinball loss~\citep{Gneiting2011Quantiles} over the forecast horizon $h$ is scaled by the average absolute error of the one-step-ahead in-sample na\"ive forecast within the period between the first non-zero sales to time $T$, and $\mathbf{1}$ is the indicator function. For example, for the 10th and 90th percentile forecast evaluation, $u \in \{0.1, 0.9\}$, and $q=2$ corresponds to the number of quantiles of interest. The WSPL is computed by the weighted average of the average SPL for all the quantiles per series with weights $w_i$,
$${\rm WSPL} = \sum_{i=1}^{n}w_i \times \frac{1}{q}\sum_{j=1}^{q}{\rm SPL}_i[u_j].$$
Each series is weighted equally in our experiments, that is, $w_i = 1/n$ and $n$ denotes the total number of series at the lower level L. A lower WSPL indicates a better estimate of the forecast intervals. 
However, when a series only has sales on the last consecutive several timestamps with the same (non-zero) amount, the scaling part of the metric in the denominator becomes zero and the error calculated is invalid. As this rarely happens, for example, 8 series with such property are present in the Favorita dataset, we omit such series during the evaluation process.

In experiments where only point forecasts are evaluated, we use the Mean Squared Error (MSE) for evaluation. Given the point forecasts $\hat{y}_t$, the MSE is defined as follows,
$${\rm MSE} =  \sum_{i=1}^{n}\frac{1}{n}  \sum_{t=T+1}^{T+h}\frac{1}{h} \left(y_t-\hat{y}_t\right)^2.$$

\section{Results and discussion}
\label{sec:evaluation}

In the following, we present an evaluation on three different datasets separately. The proposed top-down forecasting framework and sampling approach are first evaluated on the e-commerce dataset. Based on the results, we aim at transferring the findings to the brick-and-motar datasets. Therefore, we use the most competitive models for further experiments on the Favorita and M5 datasets. For the M5 dataset, we are able to directly compare the performance of the proposed top-down forecasting framework on the two layers considered with the results of the original competition participants.

\subsection{Evaluation with the e-commerce dataset}
In this section, we present detailed performance evaluations on the proprietary e-commerce dataset.

\subsubsection{Top-down forecasting approach with the two-layer hierarchy}

We use a subsample of the dataset due to the feasibility of training direct quantile models for comparison purposes. On the aggregated level A, we randomly select 500 erratic series, 2000 lumpy series, and 3000 intermittent series, and use all the smooth series (425 in total) in our experiments. This leads to a subset of altogether 17,926 series on level L, which contains 3,076 smooth series, 4,681 erratic series, 5,076 lumpy series, and 5,093 intermittent series, respectively, on this level. Models are globally trained within each category and a top-down approach is then applied. 

Table~\ref{tab:wspl-top-down-tokopedia} presents the WSPL results on level L, based on the demand classification category of the respective level A series. The benchmarks are placed at the top of the table, and models trained in a top-down fashion are arranged by distribution assumptions. Noticeably, the direct LightGBM model outperforms all other models in all categories except being in second place for smooth and intermittent data. It is somewhat surprising to find that simply using the in-sample quantiles can lead to a competitive forecasting accuracy, especially for the lumpy and intermittent series. Besides, DeepAR models beat other methods for smooth and intermittent data, while no consistent performance can be found in other categories. The LightGBM models perform well in all categories, and models implemented with a negative binomial loss function achieve the best results on smooth and erratic data. However, not much variation can be seen among the different parameterisations. Linear models are also found to be competitive as PR models and Lasso models present satisfactory results across all data categories. 
Moreover, GAMs failed to find proper coefficient sets for lumpy and intermittent series, so only two errors are reported under each distribution assumption.

With regard to different distribution assumptions, we can find that models with negative binomial assumptions outperform those with Poisson assumptions, indicating that the data is over-dispersed.

\begin{table}[H]
\fontsize{9}{9}\selectfont
  \centering
  \caption{The WSPL on level L categorised based on the demand class on level A, where top-down forecasting methods are sorted by distribution assumptions.
  }
  
    \begin{tabular}{rrrrr}
    \toprule
    Model & \multicolumn{1}{l}{Smooth} & \multicolumn{1}{l}{Erratic} & \multicolumn{1}{l}{Lumpy} & \multicolumn{1}{l}{Intermittent} \\
    \midrule
    ARIMA & 0.3595 & 0.4417 & 0.5279 & 0.6040 \\
    Drift & 1.2736 & 1.5289 & 1.8734 & 2.0144 \\
    ETS   & 0.3698 & 0.4573 & 0.5543 & 0.5540 \\
    Mean  & 0.3566 & 0.4310 & 0.4899 & 0.5152 \\
    Naive & 1.2326 & 1.4826 & 1.8173 & 1.9472 \\
    Snaive & 0.6094 & 0.7449 & 0.8858 & 0.9263 \\
    In-sample quantiles & 0.2293 & 0.2378 & 0.2192 & 0.1801 \\
    DeepAR & \textbf{0.1953} & 0.3451 & 0.2669 & \textbf{0.1724} \\
    Direct& 0.1962 & \textbf{0.2068} & \textbf{0.2028} & 0.1742 \\
    \hline
    \multicolumn{5}{l}{\textbf{Negative binomial distribution assumption}}\\[2pt]
    GAM & 0.2169 & 0.2360 & - & - \\
    Lasso & 0.2137 & 0.2301 & 0.2189 & 0.1793 \\
    
    Pooled Regression & 0.2140 & 0.2252 & 0.2115 & 0.1741 \\
    LightGBM Huber loss default & 0.2146 & 0.2255 & 0.2159 & 0.1782 \\
    LightGBM Huber loss linear leaf & 0.2145 & 0.2257 & 0.2157 & 0.1784 \\
    LightGBM L1 loss default & 0.2132 & 0.2271 & 0.2231 & 0.1816 \\
    LightGBM L1 loss linear leaf & 0.2132 & 0.2265 & 0.2382 & 0.1771 \\
    LightGBM L1 loss preset & 0.2133 & 0.2276 & 0.2210 & 0.1802 \\
    LightGBM L2 loss default & 0.2144 & 0.2280 & 0.2143 & 0.1798 \\
    LightGBM L2 loss linear leaf & 0.2155 & 0.2285 & 0.2159 & 0.1781 \\
    LightGBM L2 loss preset & 0.2168 & 0.2286 & 0.2222 & 0.1815 \\
    LightGBM Neg. Bin. loss default & 0.2138 & 0.2228 & 0.2265 & 0.1925 \\
    LightGBM Poisson loss default & 0.2160 & 0.2293 & 0.2143 & 0.1826 \\
    LightGBM Poisson loss linear leaf & 0.2158 & 0.2284 & 0.2140 & 0.1788 \\
    LightGBM Poisson loss preset & 0.2152 & 0.2271 & 0.2194 & 0.1818 \\
    LightGBM Tweedie loss default & 0.2152 & 0.2272 & 0.2149 & 0.1796 \\
    LightGBM Tweedie loss linear leaf & 0.2155 & 0.2276 & 0.2166 & 0.1779 \\    
    \hline
    \multicolumn{5}{l}{\textbf{Poisson distribution assumption}}\\[2pt]
    GAM & 0.2346 & 0.2708 & - & -\\
    Lasso & 0.2326 & 0.2564 & 0.2370 & 0.1876 \\
    
    Pooled Regression & 0.2305 & 0.2471 & 0.2201 & 0.1732 \\
    LightGBM Huber loss default & 0.2316 & 0.2469 & 0.2180 & 0.1813 \\
    LightGBM Huber loss linear leaf & 0.2321 & 0.2484 & 0.2188 & 0.1824 \\
    LightGBM L1 loss default & 0.2299 & 0.2429 & 0.2250 & 0.1795 \\
    LightGBM L1 loss linear leaf & 0.2298 & 0.2441 & 8.3141 & 0.1785 \\
    LightGBM L1 loss preset & 0.2305 & 0.2441 & 0.2205 & 0.1769 \\
    LightGBM L2 loss default & 0.2318 & 0.2533 & 0.2331 & 0.1860 \\
    LightGBM L2 loss linear leaf & 0.2325 & 0.2558 & 0.2362 & 0.1872 \\
    LightGBM L2 loss preset & 0.2340 & 0.2552 & 0.2452 & 0.1903 \\
    LightGBM Neg. Bin. loss default & 0.2241 & 0.2398 & 0.2846 & 0.5504 \\
    LightGBM Poisson loss default & 0.2340 & 0.2568 & 0.2343 & 0.1901 \\
    LightGBM Poisson loss linear leaf & 0.2339 & 0.2569 & 50.0288 & 0.1871 \\
    LightGBM Poisson loss preset & 0.2327 & 0.2519 & 0.2407 & 0.1889 \\
    LightGBM Tweedie loss default & 0.2321 & 0.2523 & 0.2322 & 0.1843 \\
    LightGBM Tweedie loss linear leaf & 0.2328 & 0.2541 & 0.2357 & 0.1815 \\
    \bottomrule
    \end{tabular}
    
  \label{tab:wspl-top-down-tokopedia}
\end{table}

We further analyse the accuracy on level A, that is, the level where forecasts are produced. With point forecasts obtained, we can estimate distribution parameters and probabilistic forecasts can be generated afterwards. For a clearer illustration, we present LightGBM models only under default parameter settings as the simplest representation. 

From Table~\ref{tab:wspl-top-down-SKU-tokopedia}, we see that the in-sample quantile forecasts perform well among all the categories. In these cases, a Poisson distributional assumption tends to be more appropriate except for the lumpy series. Regarding the proposed methods, the LightGBM models with negative binomial loss come to the top for smooth and lumpy categories, however, for the other two categories the results are the opposite, which is aligned with the performance on level L. Linear methods are still competitive on this level.

\begin{table}[htbp]
  \centering
  \caption{The WSPL evaluation on level A by category.}
   
    \begin{tabular}{rrrrr}
    \toprule
    Model & Smooth &Erratic & Lumpy &Intermittent \\
    \midrule
    In-sample quantiles & 0.2877 & 0.3139 & 0.2429 & 0.1910 \\
    \hline
    \multicolumn{5}{l}{\textbf{Negative binomial distribution assumption}}\\
    Lasso & 0.4713 & 0.4140 & 0.2586 & 0.2028 \\
    
    Pooled Regression & 0.4676 & 0.3927 & 0.2459 & 0.1957 \\
    LightGBM Neg. Bin. loss default & 0.4380 & 0.3833 & 0.2859 & 0.2199 \\
    LightGBM Poisson loss default & 0.4907 & 0.4301 & 0.2637 & 0.2094 \\
    LightGBM Tweedie loss default & 0.4802 & 0.4189 & 0.2640 & 0.2055 \\
    \hline
    \multicolumn{5}{l}{\textbf{Poisson distribution assumption}}\\
    Lasso & 0.3007 & 0.3971 & 0.2756 & 0.2010 \\
   
    Pooled Regression & 0.2821 & 0.3220 & \textbf{0.2423} & \textbf{0.1784} \\
    LightGBM Neg. Bin. loss default & \textbf{0.2483} & \textbf{0.2827} & 0.3466 & 0.6825 \\
    LightGBM Poisson loss default & 0.2970 & 0.3672 & 0.2617 & 0.2019 \\
    LightGBM Tweedie loss default & 0.2895 & 0.3479 & 0.2572 & 0.1930 \\
    \bottomrule
    \end{tabular}
    
   \label{tab:wspl-top-down-SKU-tokopedia}
\end{table}

Table~\ref{tab:tabTrainTime} compares the total training time of the forecasting models. Models were trained on a Google Cloud Platform (GCP) n1-standard-16 machine (16 vCPUs, 60 GB RAM) using R 3.6. The direct modelling approach takes up to 1,900 times longer to train than the distribution assumption approach. The overall training process of LightGBM models and linear models is fast. Modelling with GAMs involves much longer training effort and memory, and for lumpy and intermittent series, errors even occur with memory allocation when fitting the model. Within the LightGBM model variants, the ones using negative binomial loss take the longest time. This is due to the iterative optimisation of parameter $n$ of the negative binomial distribution. Finally, DeepAR models are more computationally expensive than LightGBM variants and linear models.

\begin{table}[H]
\caption{\label{tab:tabTrainTime}Training time (in minutes) for the Direct and DeepAR models on level L, and LightGBM models and linear models on level A. 
}
\centering
\begin{tabular}[t]{rrrrr}
\toprule
Model & Smooth & Erratic & Lumpy & Intermittent \\
\midrule
    Direct & 1147.20 & 5208.60 & 6049.20 & 5286.60 \\
    DeepAR & 14.9  & 15.1  & 21.54 & 22.04 \\
    \hline
    GAM   & 1814.40 & 2462.40 & -     & - \\
    Lasso & 0.45  & 0.87  & 4.46  & 6.98 \\
    
    Pooled Regression & 0.81  & 1.29  & 6.18  & 7.14 \\
    LightGBM Huber loss default & 0.39  & 0.75  & 8.62  & 16.23 \\
    LightGBM L1 loss default & 0.43  & 0.65  & 8.15  & 13.33 \\
    LightGBM L2 loss default & 0.41  & 0.66  & 7.78  & 14.09 \\
    LightGBM Neg. Bin. loss default & 1.77  & 2.03  & 9.59  & 25.26 \\
    LightGBM Poisson loss default & 0.43  & 0.75  & 8.60  & 16.23 \\
    LightGBM Tweedie loss default & 0.39  & 0.71  & 8.88  & 12.25 \\
    LightGBM L1 loss preset & 1.87  & 2.87  & 20.95 & 33.97 \\
    LightGBM L2 loss preset & 1.72  & 2.68  & 20.23 & 33.39 \\
    LightGBM Poisson loss preset & 1.97  & 2.83  & 20.95 & 33.97 \\
    LightGBM Huber loss linear leaf & 0.39  & 0.61  & 6.89  & 11.23 \\
    LightGBM L1 loss linear leaf & 0.36  & 0.61  & 7.00  & 11.52 \\
    LightGBM L2 loss linear leaf & 0.37  & 0.61  & 7.02  & 12.14 \\
    LightGBM Poisson loss linear leaf & 0.36  & 0.62  & 6.93  & 11.17 \\
    LightGBM Tweedie loss linear leaf & 0.38  & 0.61  & 6.89  & 11.47 \\
\bottomrule
\end{tabular}
\end{table}

\subsubsection{Sampling on level L with linear regression}

\begin{figure}[H]
\includegraphics[width=0.99\linewidth,]{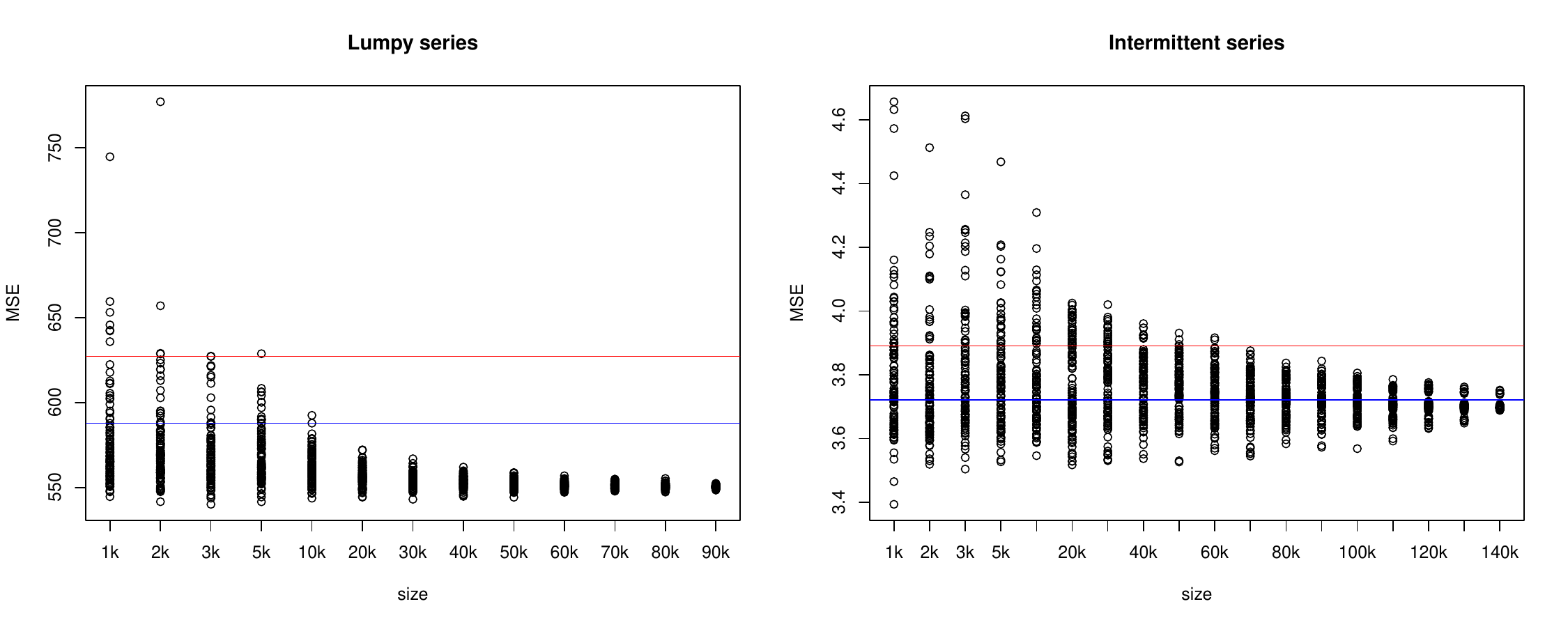} \caption{The MSE over all series w.r.t.\@ training sample size for the intermittent and lumpy category of the e-commerce dataset on level L, where the blue line and the red line represent the MSE for mean forecasts and zero forecasts, respectively.} \label{fig:lower-series-PID}
\end{figure}

In this section, we analyse how forecast accuracy evolves w.r.t.\@ sample sizes for intermittent and lumpy series, when training directly on level L, but with randomly subsampled training sets of the original dataset. Figure~\ref{fig:lower-series-PID} reports MSE of point forecasts in a 28-day horizon under 100 times repeated sampling at each sample size. We can clearly see a plateau in both categories. For lumpy series, sampling 20,000 series could be sufficient to represent the whole category. However, when comparing the accuracy with mean forecasts and all-zero forecasts, the linear model for the intermittent series is not competitive against the mean benchmark. A possible explanation is that the e-commerce series are very intermittent. Since the percentage of zeros for this category is over 95\%, we may assume that the series are very hard to predict.

\subsection{Evaluation with Kaggle Favorita dataset}
\label{sec:evaluation-favorita}
Based on the previous experiments, we limit our experiments on this dataset on a selection of the best-performing methods from the previous experiments, from the different categories of methods, to run with the Kaggle Favorita dataset, namely LightGBM with Poisson loss, Tweedie loss and negative binomial loss functions, pooled regression, and Lasso. Again, we use 100 lags as input, and LightGBM models are trained under default parameter settings. In the top-down probabilistic experiments, we assume sales data to follow a Poisson distribution or a negative binomial distribution across the hierarchy.

\subsubsection{Top-down forecasting approach}

Table~\ref{tab:favorita-wspl} reports the WSPL errors that are calculated on both aggregated level and the lower level. 
From the second column, we can compare the top-down approach against in-sample quantile forecasts. We observe that our methods are competitive, and the linear models have remarkably outperformed the LightGBM variants. From the third column, we can also generate probabilistic forecasts on the level on which the models are trained, in which case the forecasts are much more accurate compared with the in-sample quantiles. Moreover, distribution assumptions perform differently on the two layers. Even though the Poisson distribution seems to be more appropriate on the lower level, Lasso and Pooled Regression still achieve high accuracy even with a negative binomial assumption. Table~\ref{tab:favorita-train-time} reports the training time for each method. Overall, linear models and LightGBM models can be obtained at a fast speed. The Lasso model is slight slower due to fitting additional regularisation parameters. Also, the LightGBM model with negative binomial loss requires more training effort, as is to be expected.

\begin{table}[H]
  \centering
  \caption{The WSPL for lower level and the aggregated level from Favorita dataset.}
    
    \begin{tabular}{rrr}
    \toprule
    Model & Lower level &Aggregated level \\
    \midrule
    In-sample quantiles & 0.2999 & 0.5020 \\
    \hline
    \multicolumn{3}{l}{\textbf{Negative binomial distribution assumption}}\\
    Lasso & 0.2895 & 0.3690 \\
    Pooled Regression & 0.2904 & \textbf{0.3303} \\
    LightGBM Neg. Bin. loss default & 0.3028 & 0.3735 \\
    LightGBM Poisson loss default & 0.3034 & 0.3903 \\
    LightGBM Tweedie loss default & 0.3015 & 0.3781 \\
    \hline
    \multicolumn{3}{l}{\textbf{Poisson distribution assumption}}\\
    Lasso & 0.2890 & 0.5117 \\
    Pooled Regression & \textbf{0.2747} & 0.3631 \\
    LightGBM Neg. Bin. loss default & 0.2896 & 0.4199 \\
    LightGBM Poisson loss default & 0.2940 & 0.4435 \\
    LightGBM Tweedie loss default & 0.2899 & 0.4214 \\
    \bottomrule
    \end{tabular}
    
  \label{tab:favorita-wspl}
\end{table}
\begin{table}[H]
\fontsize{9}{9}\selectfont
\caption{\label{tab:favorita-train-time}Training time (in minutes) on aggregated level of Favorita dataset.}
\centering
\begin{tabular}[t]{lcccccc}
\toprule
Method & LightGBM Poisson loss & LightGBM Tweedie loss & LightGBM Neg. Bin. loss& PR & Lasso\\
\midrule
Training Time & 2.60 & 2.61 & 95.89 & 2.51 & 3.52\\
\bottomrule
\end{tabular}
\end{table}

\subsubsection{Sampling on lower level with linear regression}

\begin{figure}[H]
\includegraphics[width=0.99\linewidth,]{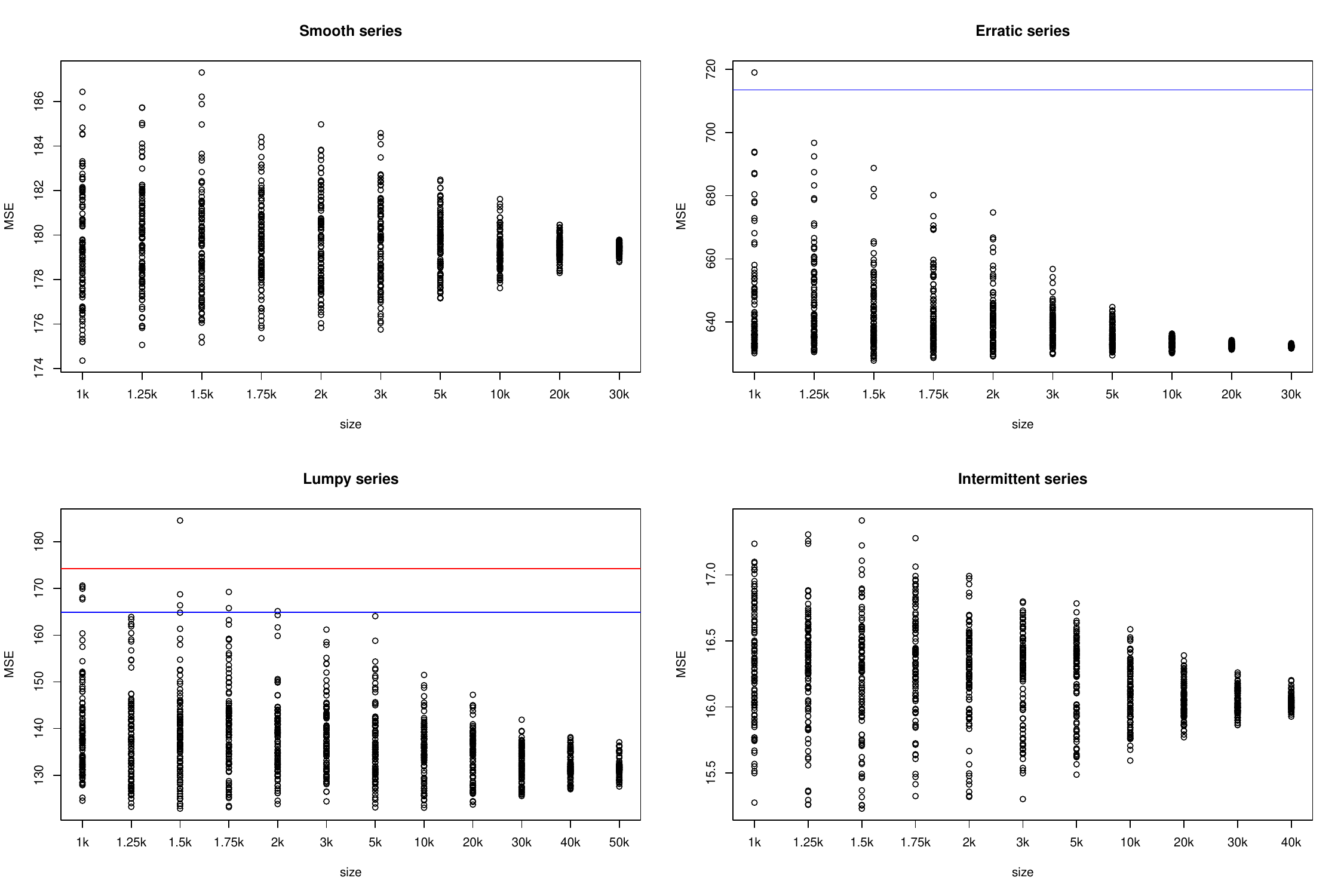} \caption{The MSE over all series w.r.t. training sample size by category of the Favorita dataset, where the blue line and the red line represent the MSE for mean forecasts and zero forecasts, respectively. When the blue and red lines are missing from the plots, it means they have errors that are too large to be displayed with the scaling in the plot, so that all points shown in the plot lie below both the red and blue lines.}\label{fig:favorita-lower-sampling}
\end{figure}

Figure~\ref{fig:favorita-lower-sampling} presents the sampling results, where we see that the linear models outperform the benchmarks of mean forecasts and zero forecasts for all categories. Besides, we see that the error stabilises with growing dataset size. Specifically, the MSE stabilises the fastest for the intermittent category. For smooth and erratic series, one can find the error plateaus around the size of 5,000, which means it is sufficient to train with a subsample of this size, instead of the whole dataset. Lumpy series take most of the proportion at the lower level, and a proper estimate of the whole category takes around 10,000 series. For intermittent series, the linear models demonstrate a much better performance compared with previous results on the e-commerce dataset, and an accuracy convergence could be found when sampling with around 10,000 series.

\subsection{The M5 competition revisit}
\label{sec:evaluation-m5}

\begin{figure}[H]
\includegraphics[width=0.99\linewidth,]{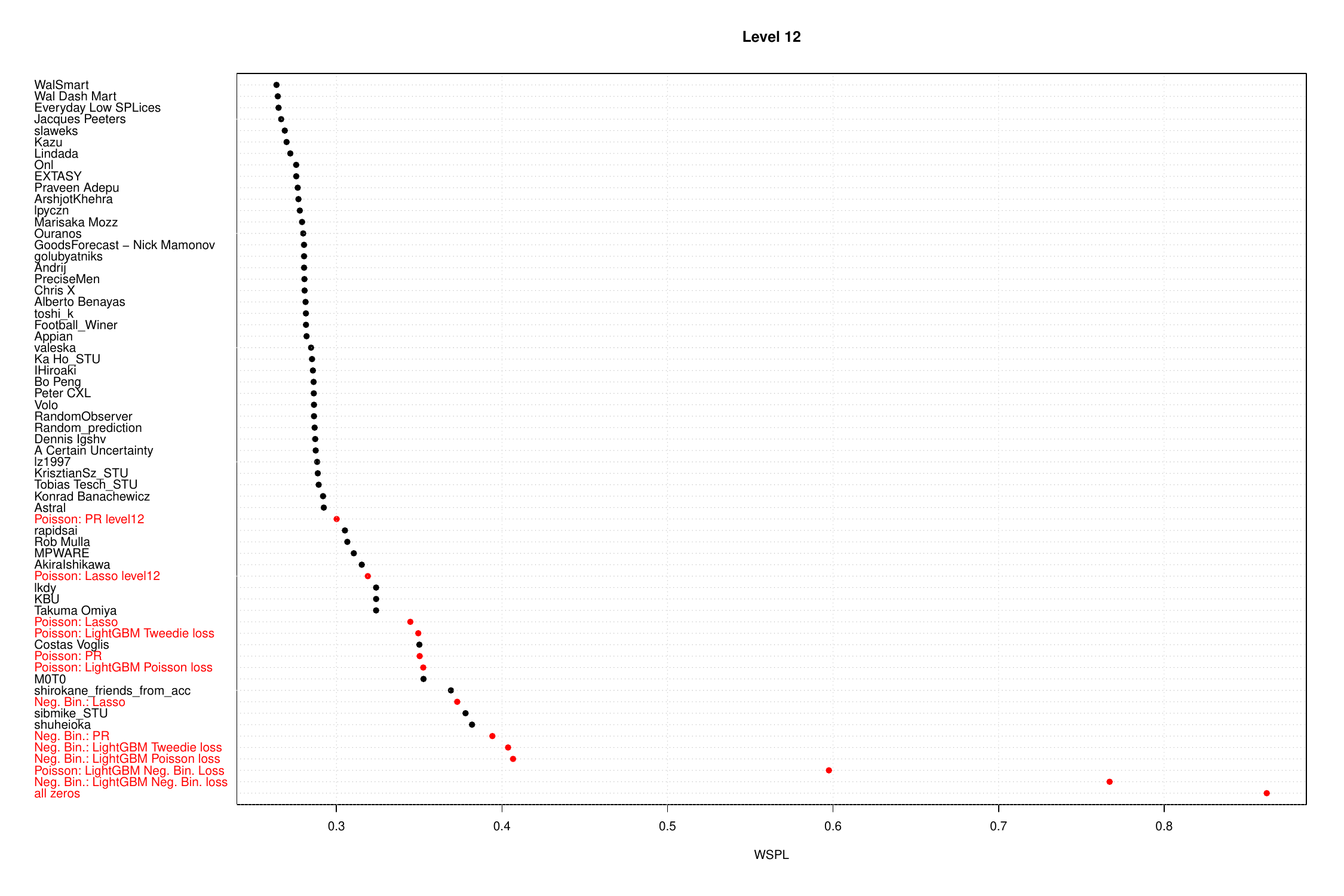} \caption{The performance of the proposed methods on level 12 compared with the top 50 submissions of the M5 uncertainty competition.} \label{fig:m5-result}
\end{figure}

We conduct experiments on the M5 dataset similar to Section~\ref{sec:evaluation-favorita} with selected models, and probabilistic forecasts are generated based on Poisson distribution or negative binomial distribution.
The names of the methods discussed in the following have the name of the specific distribution, i.e., Poisson, Neg.\@ Bin., at the beginning. Figure~\ref{fig:m5-result} compares the WSPL values on level 12 with the top 50 participants in the uncertainty track of the original competition. Remarkably, the proposed top-down forecasting approaches with Poisson distributional assumption all enter the top 50 when compared with the original 892 teams participating teams, w.r.t.\@ WSPL. We also notice that methods which assume future sales to follow a negative binomial distribution perform worse, which is not the case in the previous experiments. As the length of historical time series in the M5 dataset is over 5 years, which is much longer than the series we evaluated in the previous sections, it is reasonable to assume that the short history before has resulted in a poor estimate of future distributional parameters. For data from level 12, even though the number of series is relatively small on this level, the long history of the M5 dataset still leads to huge matrices after lag embedding, rendering many algorithms computationally infeasible. Thus, also here we perform an experiment with subsampling. Different to the earlier experiments with a simple subsampling approach, where our aim was to show which dataset size is large enough to achieve comparable results to the use of the full dataset, we here want to make full use of the dataset due to the smaller amount of series. Thus, we opt for an ensembling approach as follows. We partition the data into five disjoint folds, then train models on each fold, and finally obtain forecasts by a simple average of the forecasts obtained from each model. Within this approach, we consider two linear models trained on level 12 with Poisson distributional assumption. We then compare the accuracy and computation results with the models trained on level 10. 
From Figure~\ref{fig:m5-result}, the PR and Lasso models from level 12 ranked top compared with models train from level 10. Table~\ref{tab:m5-train-time} provides the training duration of models on level 10 and directly on level 12. The GBTs can still be trained with modest computational effort, as well as the linear models. However, even though it may provide slight accuracy gains, training directly on level 12 can take much more computational time and memory. 

\begin{figure}[H]
\includegraphics[width=0.99\linewidth,]{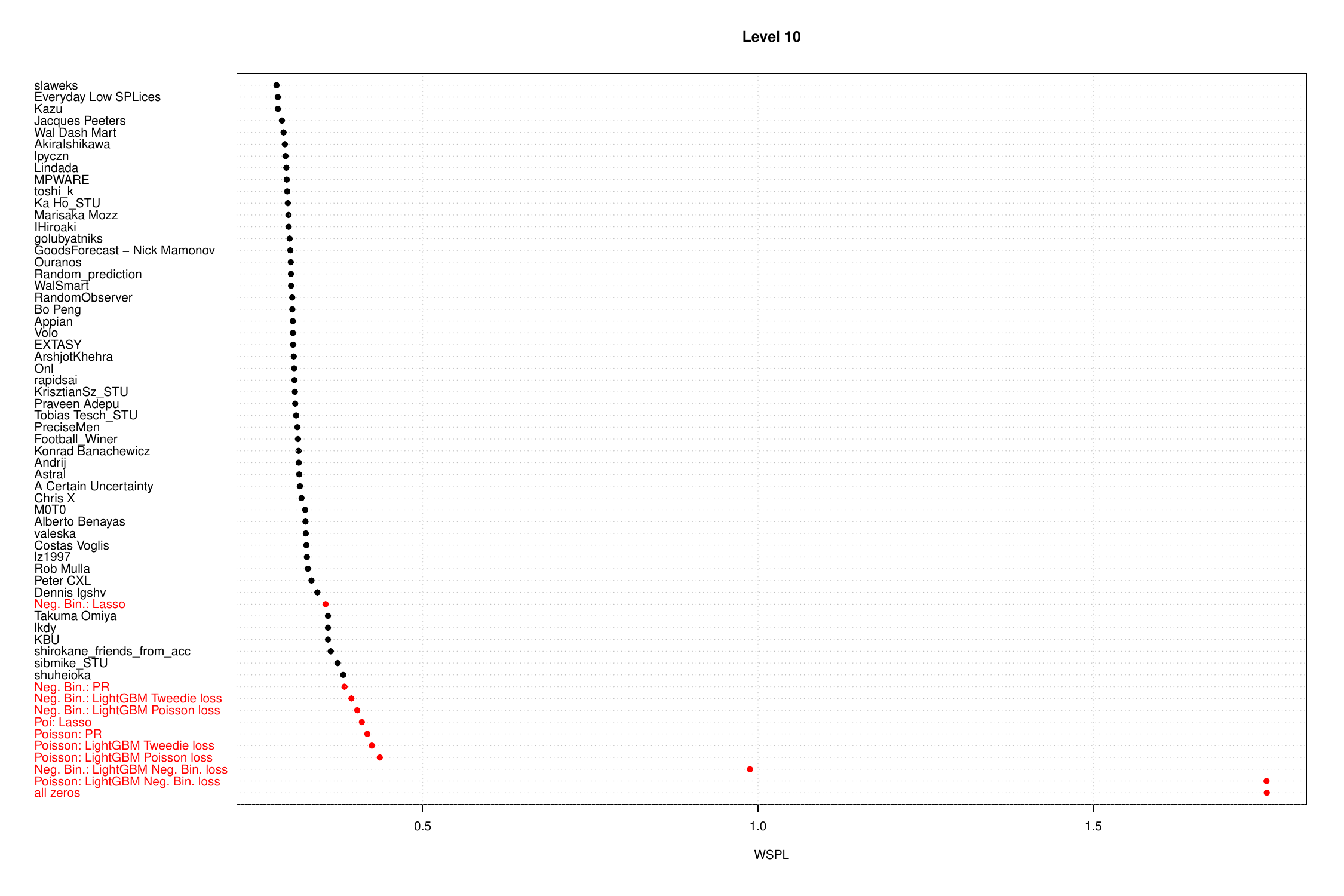} \caption{The performance of the proposed methods on level 10 compared with the top 50 submissions of the M5 uncertainty competition (the ranks of the models in the competition which are below 50 are not publicly available).} \label{fig:m5-result-level10}
\end{figure}

In Figure~\ref{fig:m5-result-level10} we present a further evaluation comparing with the top 50 entries of the M5 on level 10, the level on which we train the top-down models. On this level, the negative binomial assumption seems to be more appropriate. The Lasso model with this assumption enters the top 50, and other models with the same assumption also achieve competitive accuracy. We note that as only the results of the top 50 teams in the competition are publicly available, we are not able to obtain specific ranks of the proposed methods that rank below the top 50. Also, it is worth pointing out that the forecasts we generate are consistent within the hierarchy, while the participants were not required to consider reconciliation during the competition. However, the LightGBM models with negative binomial loss are not competitive on both levels.

\begin{table}[H]
\caption{\label{tab:m5-train-time}Training time (in minutes) of the proposed model variants on level 10 and direct modelling on level 12, with 100 lags as input.}
\centering
\resizebox{\linewidth}{!}{
\begin{tabular}[t]{lccccccc}
\toprule
Method & LightGBM Poisson loss default & LightGBM Tweedie loss default & LightGBM Neg. Bin. loss default & PR &  Lasso & PR (level 12) & Lasso (level 12)\\
\midrule
Training Time & 2.17 & 2.06  & 68.90  & 2.56  & 3.36  & 26.03  & 66.86  \\
\bottomrule
\end{tabular}}
\end{table}

\subsection{Discussion}
\label{sec:discussion}
As evaluated in the experiments, the proposed top-down forecasting framework has demonstrated the capability of producing accurate probabilistic forecasts with adequate training effort. However, we also found that the accuracy depends largely on the estimation of distributional parameters. The Poisson assumption tends to be robust in most cases, while the negative binomial distribution would require a more precise estimate of the variance. When the length of available history in the data is long, we observe that oftentimes then the variance of the overall series may not provide a valid description for future data, presumably due to structural breaks in the series. In this situation, it is beneficial to use only the most recent data for variance estimation. 

\begin{figure}[H]
\includegraphics[width=0.99\linewidth,]{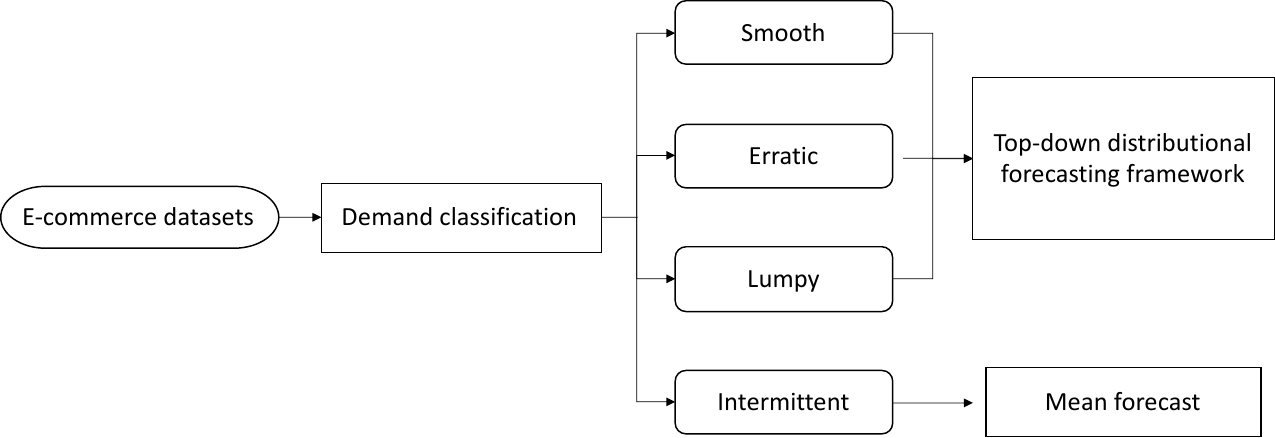} \caption{The workflow for e-commerce datasets.} \label{fig:workflow}
\end{figure}

It is an interesting finding that the linear model is not competitive against a mean forecast in the intermittent series of the e-commerce data, however, this is not the case in the Favorita dataset and the M5 dataset, where brick-and-motar sales data is considered. 
Recall the percentage of zeros calculated in Table~\ref{tab:data-summary}  on the lower level of the three datasets analysed in this research. It is noticeable that in the intermittent series of the e-commerce data over 98\% of entries are zero, implying a high degree of intermittency. This explains why these series are relatively unpredictable and no method leads any benefits over the most simple benchmarks. Taking all the findings into account, we summarise a general workflow for our e-commerce forecasting use case in Figure~\ref{fig:workflow}.

\section{Conclusion}
\label{sec:conclusion}

In this paper, we have proposed a scalable forecasting framework which is capable of generating reliable probabilistic forecasts at a fast speed. Direct modelling on the lower level and producing quantile forecasts is accurate, but the computational effort is large while no corresponding large gains accuracy are observed. In our use cases, and presumably many others in the industry, the additional computational effort is thus not justified. 
Our forecasting system is feasible to implement in production. The top-down forecasting framework has also been evaluated with two public datasets and has shown good results. Besides, subsamples at the lower level after demand classification can be seen as a proper representation of the whole category, on which models could be trained in a feasible way. 

Probabilistic forecasts are generated based on distributional assumptions. Since the variance of sales is often larger than the mean, it is more appropriate to assume a negative binomial distribution than a Poisson distribution. However, the performance of the former one is more sensitive to parameter estimation. Moreover, we have implemented the negative binomial loss function in the common GBT package LightGBM. Based on our experiments, we argue that it is also reasonable to make a Poisson distribution assumption with GBTs which is still able to obtain good forecast accuracy. 

Somewhat surprisingly, linear models are competitive with the state-of-the-art LightGBM algorithm in situations where no external covariates are used (as in our research; external variables could regard pricing, promotions, and others). Here, linear models offer a simple alternative to GBTs that is fast, robust, and more interpretable.

For future directions, as the distributional forecasts are based on parameter estimation, improving point forecasts on the aggregated level will improve the overall accuracy. The proposed framework is using the total historical proportions during disaggregation, since this is a static top-down approach, using a disaggregation method that accounts for future changes may also improve forecasting accuracy.

\section*{Acknowledgements}

Christoph Bergmeir is supported by a María Zambrano (Senior) Fellowship that is funded by the Spanish Ministry of Universities and Next Generation funds from the European Union.

\appendix
\section{Implementation of negative binomial loss function with LightGBM}
\label{appendix:a}

As sales data is usually over-dispersed, i.e., the variance is greater than its mean, when we use machine learning algorithms to predict the future mean values, it is a natural choice to consider the negative binomial loss function for model training. However, the LightGBM package~\citep{ke2017lightgbm} does not provide a built-in negative binomial loss function, but it provides functionality which supports user-defined loss functions. 

In order to implement any customised loss, there are two functions we need to specify: an objective function and an evaluation function. The objective function is defined according to the log likelihood of a certain distribution, and the evaluation function returns the first and second derivatives w.r.t. model predictions.

For the negative binomial distribution, the probability mass function is given by
$$P(x \vbar r, p) = {r+x-1\choose r-1}p^{r}(1-p)^{x},$$
with a mean value $\mu$ that equals to $\frac{(1-p)r}{p}$. So if we substitute $p$ w.r.t. $\mu$, that is, $p = \frac{r}{\mu+r}$, we can get the following,
$$P(x \vbar r, \mu) = {r+x-1\choose r-1} \left(\frac{r}{\mu+r}\right)^{r} \left(\frac{\mu}{\mu+r}\right)^{x}.$$
Then we rewrite the binomial coefficient with the Gamma function,
$$P(x \vbar r, \mu) = \frac{\Gamma (r+x)}{\Gamma (r) \Gamma (x+1)} \left(\frac{r}{\mu+r}\right)^{r} \left(\frac{\mu}{\mu+r}\right)^{x}.$$
So, the negative log likelihood is given by
$$L(x \vbar \mu, r) = -{\rm log} \Gamma (r+x) + {\rm log} \Gamma (r) + {\rm log} \Gamma (x+1) - r {\rm log} r + r {\rm log} (\mu+r) - x {\rm log}\mu + x {\rm log}(\mu + r).$$
And we denote the predicted mean value from the LightGBM model as $f$. As the support of the negative binomial distribution is the set of positive integers, we apply a log transformation so that $f$ is allowed to take any real value and $e^f$ is always non-negative. For data point $x_i$, treating $x_i$ as the true value and plugging in the predicted mean value after transformation, i.e., $e^{f_i}$, then the negative log likelihood is given by,
$$L(x_i \vbar f_i,r) = -{\rm log} \Gamma (r+x_i) + {\rm log} \Gamma (r) + {\rm log} \Gamma (x_i+1) - r {\rm log} r + r {\rm log} (e^{f_i}+r) - x_i f_i + x_i {\rm log} (e^{f_i} + r). $$
Consider ${\bf x} = (x_1,\ldots,x_n)$ and ${\bf f} = (f_1,\ldots,f_n)$; then our objective function is defined as
$$L({\bf x} \vbar {\bf f},r) = \sum_{i=1}^n L(x_i;f_i,r).$$
And we calculate the gradient and Hessian w.r.t. $f$,
$$g({\bf x} \vbar {\bf f},r) = \sum_{i=1}^n \left(\frac{e^{f_i} (r+x_i)}{e^{f_i}+r} - x_i \right),$$
$$h({\bf x} \vbar {\bf f},r) = \sum_{i=1}^n \frac{e^{f_i} r (r+x_i)}{(e^{f_i}+r)^2}. $$
With this we have defined all the required functions for implementation, however, the value of $r$ has to be obtained for completing the calculation. Intuitively, we can treat $r$ as a model parameter and optimise it alongside the training process, but the LightGBM package does not provide an option for defining custom parameters. A possible solution is to use coordinate-wise optimisation, that is, updating the model and $r$ iteratively until convergence. We initialise the value of $r$ by the method of moments from the historical data. The optimisation process of each iteration takes three steps: (1) train a LightGBM model with the custom loss function and the current value of $r$, (2) predict the training set with the model obtained and then get the predicted mean values, and (3) get an updated estimate of $r$ by minimising the negative log likelihood, which is also the function $L$ defined above. In this case, the LightGBM models are retrained iteratively through coordinate-wise optimisation and the optimisation procedure takes longer as the length of series grows, which in return leads to an overall longer training process.  

\bibliographystyle{elsarticle-harv}
\bibliography{references}

\end{document}